# Accuracy, Uncertainty, and Adaptability of Automatic Myocardial ASL Segmentation using Deep CNN

Hung P. Do[1,*], PhD, MSEE, Yi Guo[1], PhD, Andrew J. Yoon[2], MD, Krishna S. Nayak[1], PhD

1. *Ming Hsieh Department of Electrical and Computer Engineering, Viterbi School of Engineering, University of Southern California, Los Angeles, California, USA*
2. *Long Beach Memorial Medical Center, University of California Irvine, California, USA*

\* Corresponding author's current address and contact information:

Hung P. Do, PhD, MSEE

Canon Medical Systems USA, Inc.

2441 Michelle Drive, Tustin, CA, 92780

Phone: (714) 669-7369

E-mail: hdo@us.medical.canon

## ABSTRACT (250 words maximum)

**Purpose**: To apply deep CNN to the segmentation task in myocardial arterial spin labeled (ASL) perfusion imaging and to develop methods that measure uncertainty and that adapt the CNN model to a specific false positive vs. false negative tradeoff.

**Methods**: The Monte Carlo dropout (MCD) U-Net was trained on data from 22 subjects and tested on data from 6 heart transplant recipients. Manual segmentation and regional myocardial blood flow (MBF) were available for comparison. We consider two global uncertainty measures, named "Dice Uncertainty" and "MCD Uncertainty", which were calculated with and without the use of manual segmentation, respectively. Tversky loss function with a hyperparameter β was used to adapt the model to a specific false positive vs. false negative tradeoff.

**Results**: The MCD U-Net achieved Dice coefficient of 0.91 ± 0.04 on the test set. MBF measured using automatic segmentations was highly correlated to that measured using the manual segmentation ($R^2$ = 0.96). Dice Uncertainty and MCD Uncertainty were in good agreement ($R^2$ = 0.64). As β increased, the false positive rate systematically decreased and false negative rate systematically increased.

**Conclusion**: We demonstrate the feasibility of deep CNN for automatic segmentation of myocardial ASL, with good accuracy. We also introduce two simple methods for assessing model uncertainty. Finally, we demonstrate the ability to adapt the CNN model to a specific false positive vs. false negative tradeoff. These findings are directly relevant to automatic segmentation in quantitative cardiac MRI and are broadly applicable to automatic segmentation problems in diagnostic imaging.

# INTRODUCTION

Myocardial Arterial Spin Labeling (ASL) is a non-contrast quantitative perfusion technique that can assess coronary artery disease (1). Manual segmentation of left ventricular (LV) myocardium is a required step in the post-processing pipeline and is a major bottleneck due to the low and inconsistent signal-to-noise ratio (SNR) and blood-myocardium contrast-to-noise ratio (CNR) in the source images. More generally, segmentation of LV myocardium is a key step in the post-processing pipeline of all quantitative myocardial imaging. Segmentation masks are needed to make volumetric measurements, to provide semantic delineation of different tissues (e.g. myocardium vs. blood vs. epicardial fat), and in many cases to map measurements to a bullseye plot for convenient visualization (2).

Convolutional neural networks (CNN) have been successfully applied to automatic segmentation in several MRI applications (3–6). For example, Bai et al. recently demonstrated that CNN can provide a performance on par with human experts in analyzing cine cardiovascular magnetic resonance (CMR) data (5). Cine CMR data typically has high spatial temporal resolution, excellent SNR, and consistent blood-myocardium CNR throughout the cardiac cycle, which is why cine CMR is the gold standard for assessment of ventricular function, volumes, mass, and ejection fraction (7). In contrast to cine CMR, quantitative myocardial CMR techniques (including myocardial ASL, myocardial BOLD, myocardial first-pass perfusion, multi-parametric myocardial relaxometry, myocardial diffusion tensor imaging, etc.) (8–10) are often comprised of images with substantially lower spatial resolution, SNR, and CNR, which can vary between images due to factors such as variability in contrast preparation or heart rate. These are all reasons why automatic segmentation in quantitative CMR remains a significant challenge.

Automatic segmentation has been developed and used for many years (11,12), but the post-processing pipeline remains semi-automatic since a human operator is required to verify segmentation mask quality before commencing to the next step in the pipeline. A global score of model uncertainty is therefore desired for automatic quality control at production, for model improvement via active learning (13), and for out-of-distribution detection. Model uncertainty can be estimated using a Bayesian approach, in which not only parameters are estimated but also their posterior distributions. Dropout has been demonstrated as a Bayesian approximation, which provides model uncertainty via Monte Carlo (MC) dropout at test time (14–16). MC dropout has been used to measure model uncertainty in many segmentation problems (17–20). These studies demonstrate that pixel-wise uncertainty maps can be achieved using MC dropout at test time that allows qualitative assessment of predicted segmentations. A global quantitative score for model uncertainty, however, is desirable for automating the quality assessment, which may enable automatic post-processing pipeline.

For quantitative CMR, the AHA 17-segment model (2) is often used in the form of bullseye plot for visualization and diagnosis. To generate the bullseye plot, segmentation of left ventricular myocardium is required. Left ventricular myocardium is surrounded by ventricular blood pools and epicardial fat, which have distinct physical properties as well as spin (magnetization) history (21). Careful and conservative manual segmentation of myocardium is often required to minimize partial volume effects (22–25). Therefore, a model with a lower false positive rate may be preferred over that with a higher false positive rate. Note that false positive means pixels predicted by an algorithm that are not present in the reference segmentation (i.e. excessive segmentation).

This study aimed to (i) apply deep CNN for automatic segmentation of myocardial arterial spin labeled (ASL) data, which has low and inconsistent SNR and CNR, (ii) to measure a global score of model uncertainty without the use of the reference segmentation using MC dropout, and (iii) to adapt the network to the specific false positive and false negative needs of the application using Tversky loss function. Additionally, model uncertainty was calculated using probabilistic U-Net (26) and then compared with that calculated using Monte Carlo dropout (MCD) U-Net.

## METHODS

### Network Architecture

We implemented a CNN model based on the U-Net architecture (27) called MCD U-Net with the following modifications: (i) increased filter size from 3x3 to 5x5, (ii) added batch normalization (BN) (28) after every convolutional layer, and (iii) added dropout (15) with dropout rate of 50% at the end of every resolution scale, as seen in **Figure 1**. Similar to the original U-Net architecture, the number of base feature maps per convolutional layer in the first resolution scale was 64 that was doubled and halved in the next resolution scale in the encoding path and the decoding path, respectively. The MCD U-Net was implemented in Keras (29) with TensorFlow (30) backend.

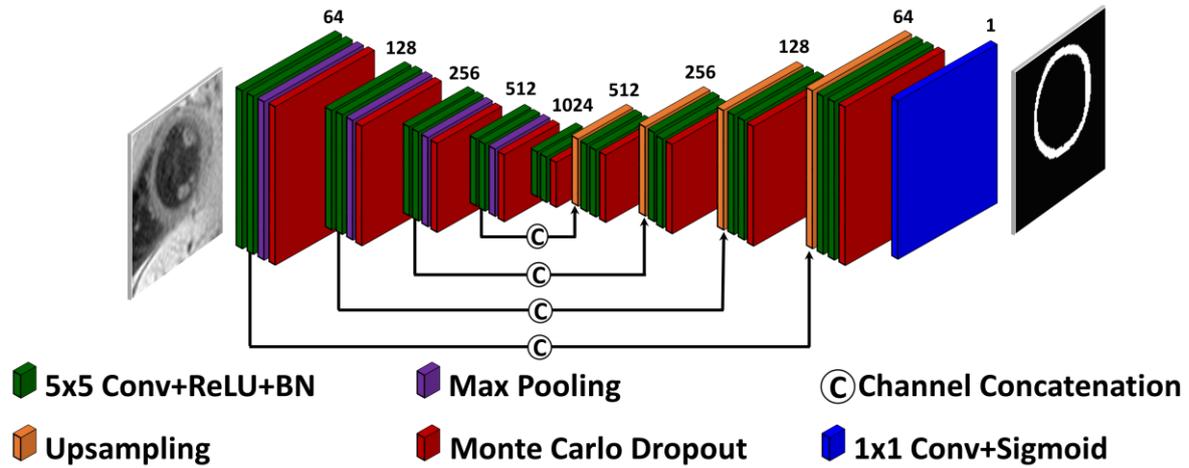

**Figure 1.** The network architecture of MCD U-Net. The modifications are that (i) 5x5 filter was used for all convolutional layers, (ii) batch normalization was added after each convolutional layer, (iii) dropout rate of 50% was added after each convolutional scale of the down- and up-sampled paths. The number of convolutional kernels is noted on top of the convolutional layers. Conv: convolutional layer; ReLU: rectified linear unit; BN: batch normalization

**Dataset**

Our dataset included 478 ASL images (control and labeled images) from 22 subjects. These were randomly divided into training and validation sets of 438 and 40 images, respectively. Trained networks were tested on 144 unseen ASL images acquired at rest and during adenosine stress from 6 heart transplant recipients. Training, validation, and testing data were drawn from previously published studies (31–33), where manual segmentation and quantitative reginal MBF from all subjects were readily available.

ASL acquisition details are summarized here, and a complete description can be found in Do et al., (31). Each ASL data set consists of six pairs of control and labeled images. Each pair of control and labeled images was acquired in a 12-second breathhold using balanced steady state free precession (bSSFP) image acquisition preceded by a flow-sensitive alternating inversion recovery (FAIR) (34,35) control (slab selective inversion) and labeled (nonselective inversion) pulse, respectively. The control/labeled pulse and the bSSFP image acquisition were cardiac triggered to occur at mid-diastole in consecutive heartbeats. bSSFP parameters: TR/TE = 3.2/1.5 ms, flip angle = $50^0$, slice thickness = 10 mm, matrix size = 96x96, and parallel acceleration factor of 2 for SENSE (36) or 1.6 for GRAPPA (37). Buxton's general kinetic model (38) was used for MBF quantification. Physiological noise (PN) was defined as a short-term variability of MBF measurement and calculated as a standard deviation of six MBF values measured from six pairs of control and labeled images (39). Dataset was normalized to have zero mean and unit variance. No data augmentation was used in training of MCD U-Net.

**Monte Carlo Dropout for Uncertainty Measure**

The concept of using Monte Carlo (MC) dropout to evaluate model uncertainty was first introduced by Hinton el al., in an online lecture (14) and manuscript (15), respectively. Model uncertainty can be calculated from a network trained with dropout rate of 50% in every (hidden) layer. At test time, multiple predictions of the same input can be obtained by running the stochastic model several times. Final prediction and model uncertainty is simply the mean and standard deviation of the stochastic predictions, respectively. Subsequently, Gal et al., (16) demonstrated that any neural network with dropout added in every weight layer is mathematically equivalent to an approximation of the Bayesian model. Hence, model uncertainty can be estimated given the posterior distribution of the trained weights.

MC dropout has been applied to evaluate model uncertainty in semantic segmentation tasks in both computer vison and medical imaging applications (20,40). In these studies, the typical output of the model uncertainty is a standard deviation pixel-by-pixel map that provides spatial information detailing where, within the image, the model is uncertain. However, a global quantitative score of model uncertainty on a specific input is desired for automatic quality assessment, for triaging images for active learning, and for out-of-distribution detection. In this study, we introduce and evaluate two global quantitative scores of model uncertainty, which are named "Dice Uncertainty" and "MCD Uncertainty". These are the measures of model uncertainty estimated with and without the use of manual segmentation, respectively.

**Tversky Loss Function for Model Adaptability**

Binary cross-entropy and Dice loss functions are often used to train a CNN model for automatic segmentation (5,41–43). The definition of the binary cross-entropy (BCE) is as follows:

$$L_{BCE} = -\frac{1}{K}\sum_{i=1}^{K}[y_i \cdot \log(\hat{y}_i) + (1-y_i) \cdot \log(1-\hat{y}_i)],$$

where K is total number of pixels in the image, $y_i$ and $\hat{y}_i$ are values of the reference and predicted mask at the $i^{th}$ pixel. Below is the definition of Dice loss function.

$$L_{Dice} = 1 - \frac{2 \cdot |A \cap B|}{|A| + |B|},$$

where A is the predicted segmentation and B is the reference segmentation. Alternatively, Dice loss function can also be defined in terms of false positive and false negative as follow.

$$L_{Dice} = 1 - \frac{TP}{TP + 0.5 \cdot FP + 0.5 \cdot FN},$$

where TP is true positive, and FP and FN are false positive and false negative, respectively. As we can see from the above equation, Dice coefficient weights false positive and false negative equally which may not be optimal for myocardial segmentation because of partial volume effects. Myocardium is surrounded by ventricular blood pools and epicardial fat, which have very different physical properties and spin (magnetization) history compared to myocardium. Therefore, false negative may be preferred over false positive. To adapt the network to the desired false positive vs. false negative tradeoff, Tversky (44) loss function could be used and is defined as below.

$$L_{Tversky} = 1 - \frac{TP}{TP + (1-\beta) \cdot FP + \beta \cdot FN},$$

where $\beta$ is a hyper-parameter that could be set during training. By adjusting $\beta$ during training, one could adapt the network to output the specific false positive vs. false negative tradeoff.

**Experiments**

All experiments were performed on a NVIDIA K80 GPU with 12 Gb RAM. Network architectures were implemented using Keras (29) with TensorFlow backend (30). Common training parameters are number of epochs = 150, batch size = 12, learning rate = 1e-4, dropout rate = 50%, and adaptive moment estimation (Adam) optimizer (45).

Accuracy

The MCD U-Net architecture was trained with Dice loss function. Training and validation loss were recorded. Dice coefficients of the test set were calculated to evaluate model accuracy. Quantitative MBF measured using automatic segmentation was compared against that measured using the reference manual segmentation using linear regression analysis.

To investigate the efficacy different training procedures, MCD U-Net was also trained on the control images only and then fine-tuned to labeled images. The Dice accuracy was then calculated and compared was that from the model trained using control and labeled images simultaneously.

Uncertainty

The MCD U-Net architecture was trained with Dice loss function and MC dropout was turned on during inference. At test time, MCD inference was applied N = 1115 times (N is called number of MC trials) on each and every test images (total 144 test images). Mean and standard deviation of 1115 predicted masks yield final predicted segmentation and pixel-by-pixel uncertainty map for each test image, respectively.

In this study, we propose two global scores of model uncertainty--"Dice Uncertainty" and "MCD Uncertainty"--which are calculated with and without the use of reference segmentation, respectively. Dice Uncertainty is defined as standard deviation of Dice coefficients calculated from N stochastic predicted segmentations given the ground truth reference segmentation. Higher Dice Uncertainty means the model is less certain of its predictions and experiences higher variability.

MCD Uncertainty provides a global score of how certain or uncertain the model is given an input image. MCD Uncertainty is defined as sum of values of all pixels in the uncertainty map normalized by the volume of the predicted mask. Similar to mean squared error (MSE) or structural similarity (SSIM) index, MCD Uncertainty is expected to provide a global view on how certain or uncertain the model is for a specific test image. That potentially allows automatic quality assessment of automatic segmentation without the needs of the reference ground truth segmentation, which is typically not available at production.

Time penalty is a major consideration when using MC dropout for model uncertainty measure (18). We carried out two experiments that studied the effects of number of MC trials and batch size on uncertainty measure and inference time, respectively. In the first experiment, we performed MC dropout on one test case with 16384 MC trials. Smaller MC trials was retrospectively bootstrapped from a distribution of 16384 samples 1024 times. That allows calculation of confidence interval (i.e. standard deviation) of uncertainty measure as a function of number of MC trials. In the second experiment, we performed N=1024 MC trials on a single test case to measure inference time, mean prediction, and Dice Uncertainty as a function of batch size.

Adaptability

The MCD U-Net architecture was trained with Tversky loss function, which has a hyper parameter β. The MCD U-Net model was trained with nine different values of β ranging from 0.1 to 0.9 with step size of 0.1. False positive and false negative rate were defined as an average number of false positive and false negative pixels per image, respectively. False positive rate and false negative rate were calculated and compared to that from the model trained with binary cross-entropy (BCE) loss and Dice loss.

To demonstrate the consequence of partial volume effects, "thin mask" and "thick mask" were generated using a "*bwmorph*" function in Matlab that removes and adds one pixel from both sides of the reference masks in the test set, respectively. Average Dice coefficient, false positive and false negative rate were calculated and compared with that calculated from CNN models. Quantitative MBF measured using "thin mask" and "thick mask" were compared against that from the reference masks to demonstrate that false positive is more detrimental than false negative.

**Probabilistic U-Net**

Probabilistic U-Net (prob U-Net) (26) has recently been proposed for segmentation of ambiguous images including medical imaging. The primary goal of the prob U-Net is to efficiently generate many (even infinite) plausible segmentation hypotheses for a given input image. It combines a U-Net architecture with a conditional variational autoencoder. MCD U-Net and prob U-Net are both probabilistic models that could generate unlimited number of plausible predictions by sampling the posterior distributions of trained weights and variational latent representation (lower-dimensional space), respectively.

Prob U-Net (46) was adapted and trained on the ASL images with the same data training, validation, and testing splits. Training details were similar to the medical imaging example in the original work (26). The training was performed with randomly initialized weights for over 5400 stochastic gradient decent iterations; the initial learning rate was $1e^{-4}$ that was lowered to $5e^{-5}$ after one third and to $1e^{-5}$ after two third of the iterations. Batch size was 12. Adam optimizer with default parameters (45) was used in combination with $1e^{-5}$ multiplier of weight decay. In addition to binary cross-entropy loss, Kullback-Leibler divergence with a multiplier $\lambda$ was added to compose the total loss. The Kullback-Leibler divergence penalizes the differences between the posterior and the prior distributions. Several values of $\lambda$ (0.1, 0.5, 0.75, 1.0, 2.0, 5.0, and 10.0) were tested and $\lambda=10$ was chosen given the minimum validation loss and Kullback-Leibler divergence. In this experiment, six-dimensional latent space was used. Random elastic deformation, random rotation were used for data augmentation.

**Data Analysis**

Accuracy

To access accuracy of the automatic segmentation method, Dice coefficients of a test set were calculated. Furthermore, MBF calculated using the automatic segmentations was compared against that calculated using the reference manual segmentations using linear regression and concordance correlation analyses.

To investigate if there is a correlation between contrast-to-noise ratio (CNR) and Dice accuracy, CNR was calculated for all images in the test set. Linear regression analysis between CNR and Dice accuracy was performed.

Uncertainty measure

For each test case, MC Dropout inference was performed N = 1115 times resulted in 1115 stochastic predictions. Mean predicted segmentation, pixel-by-pixel uncertainty map, Dice Uncertainty, and MCD Uncertainty were calculated as described above. Linear regression analysis was carried out to study relationship between Dice Uncertainty and MCD Uncertainty. Linear regression analysis was performed to determine if there is a relationship between MCD Uncertainty and physiological noise (PN) of the MBF measurement.

Adaptability

False positive and false negative rate, defined as an average number of false positive and false negative pixels per image, were calculated given the reference masks and predicted masks. False positive and false negative rate from networks trained with binary cross-entropy, Dice, and Tversky losses and that from the "thick mask" and "thin mask" were compared.

Probabilistic U-Net

For each test case, prob U-Net was inferenced N=128 times resulting in 128 stochastic predictions. Mean predicted segmentation, pixel-wise uncertainty map, Dice Uncertainty, and MCD Uncertainty associated with prob U-Net were calculated in the same manner as those for MCD U-Net. Dice accuracy associated with prob U-Net was calculated and compared with that of MCD U-Net. Linear regression analysis was carried out to study relationship between Dice Uncertainty and MCD Uncertainty calculated from prob U-Net. Furthermore, MCD Uncertainty calculated using MCD U-Net and prob U-Net were compared using linear regression analysis.

## RESULTS

**Accuracy**

The model was trained for 150 epochs. The training and validation loss were shown in **Supporting Information Figure S1**. Representative segmentation masks and MBF maps generated using the CNN model in comparison with that using manual segmentation are shown in **Figure 2**. Average Dice coefficient for the test set was 0.91 ± 0.04. For quantitative imaging, accuracy assessment using clinically relevant quantity is desired. Quantitative regional MBF measured using automatic segmentation is highly correlated with that calculated using manual segmentation ($R^2$ = 0.96) as in **Figure 3**. The concordance correlation coefficient was 0.98.

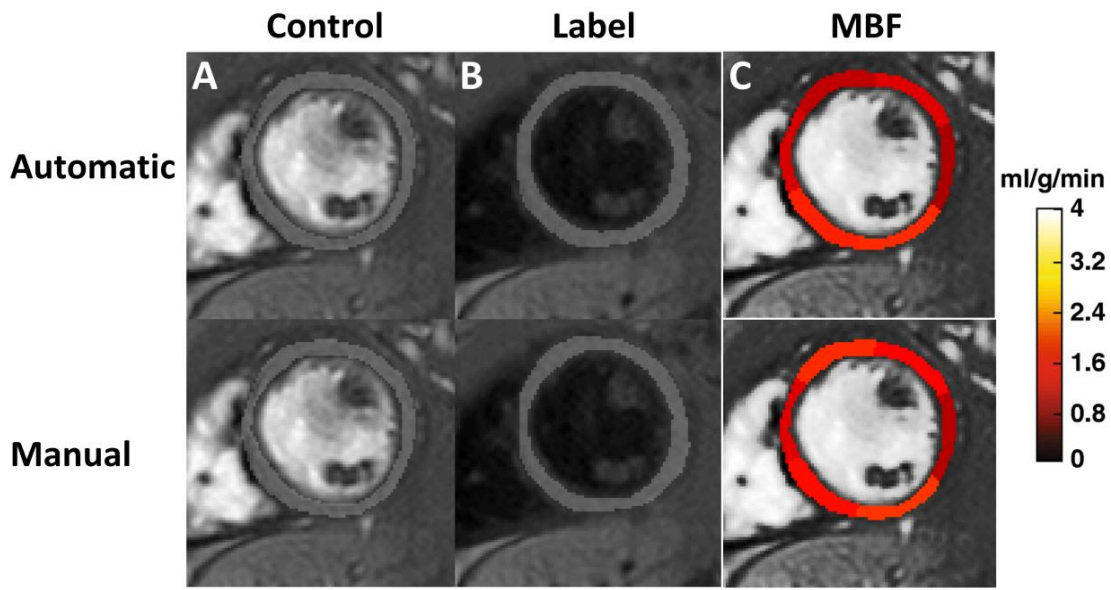

**Figure 2.** Examples of image quality, segmentation, and myocardial blood flow (MBF) maps. Shown are representative (A) control images, (B) labeled images, and (C) regional MBF maps. Segmentation masks and MBF maps generated by CNN (top row) are comparable to those from manual segmentation (bottom row).

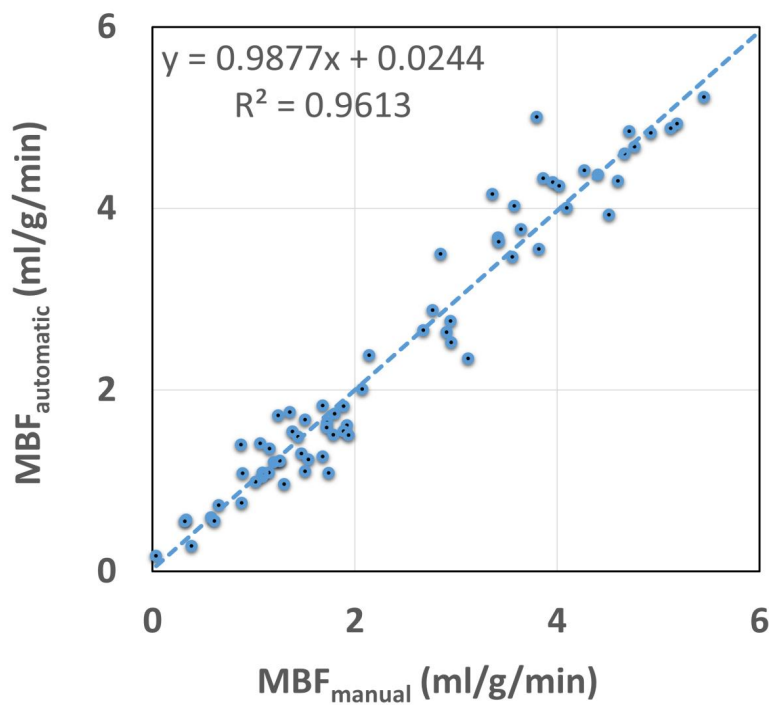

**Figure 3.** Evaluation of automatic segmentation. Regional myocardial blood flow (MBF) measured using automatic segmentation (y-axis) is highly correlated with that measured using manual segmentation (x-axis) ($R^2 = 0.96$). The concordance correlation coefficient was 0.98.

Examples of control and labeled images are shown in **Supporting Information Figure 2**. Average CNR of control and labeled images were 21.44 ± 3.81 and 6.46± 1.03, respectively. Average Dice accuracy from the control images (0.92 ± 0.02) is significantly higher than that (0.90 ± 0.04) from the labeled images (P < 0.001), however, the difference is small. There was no significant correlation between CNR and Dice accuracy ($R^2$ = 0.09).

As seen in the **Table 1**, the model trained using control and labeled images simultaneously demonstrated higher Dice accuracy compared to that obtained from the model trained on the control images and then fine-tuned on the labeled images.

**Table 1.** Comparison of Dice accuracy from the two training procedures.

|  | Pre-train with Control images and Fine-tune with the Labeled images | | | Train with both Control and Labeled images | |
|---|---|---|---|---|---|
|  | Without Fine-tuning | | Fine-tuning | | |
|  | Control | Labeled | Labeled | Control | Labeled |
| Dice Accuracy | 0.87 ± 0.04 | 0.14 ± 0.08 | 0.86 ± 0.06 | 0.92 ± 0.02 | 0.90 ± 0.04 |

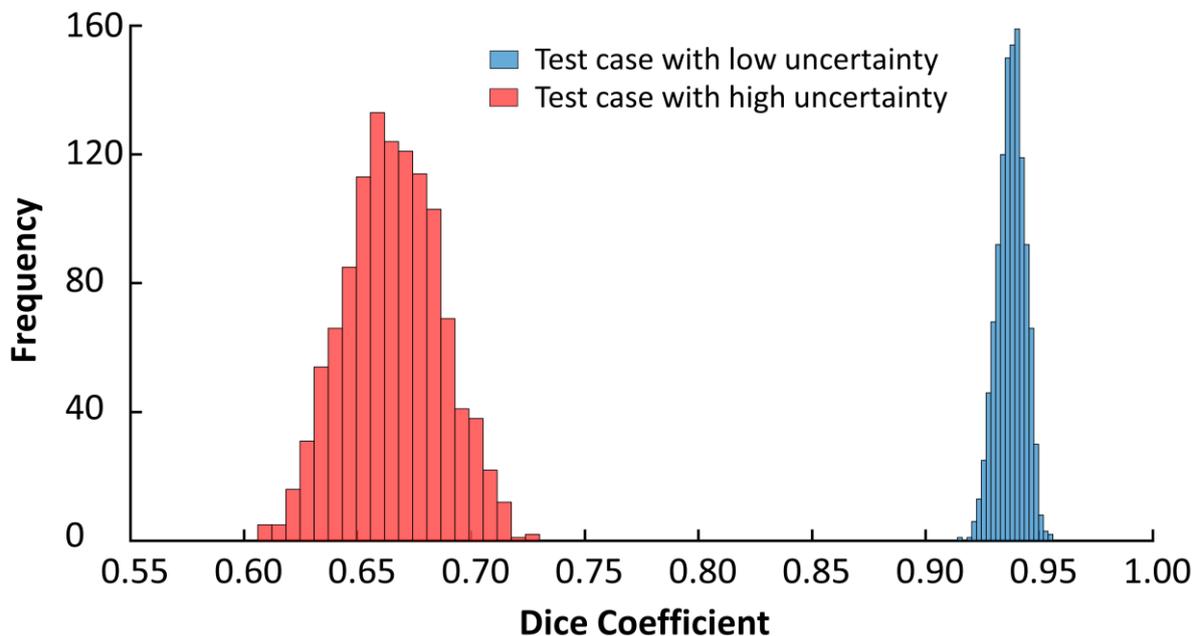

**Figure 4.** Illustration of the proposed Dice Uncertainty metric. Shown are the Dice coefficient distributions of two test cases after Monte Carlo dropout inference. The distribution is broader in the test case with high uncertainty (red) compared to that with low uncertainty (blue).

**Uncertainty**

Given the manual segmentation, the Dice coefficient distributions calculated from two test cases are shown in **Figure 4** and in the **Supporting Information Figure S3**. Dice coefficient distributions of four other test cases are shown in the **Supporting Information Figure S4**. The variance of the Dice coefficient distribution represents how the stochastic predictions fluctuate. Therefore, we proposed to use standard deviation (named Dice Uncertainty) of the Dice coefficient distribution as a measure of model uncertainty given the manual segmentation.

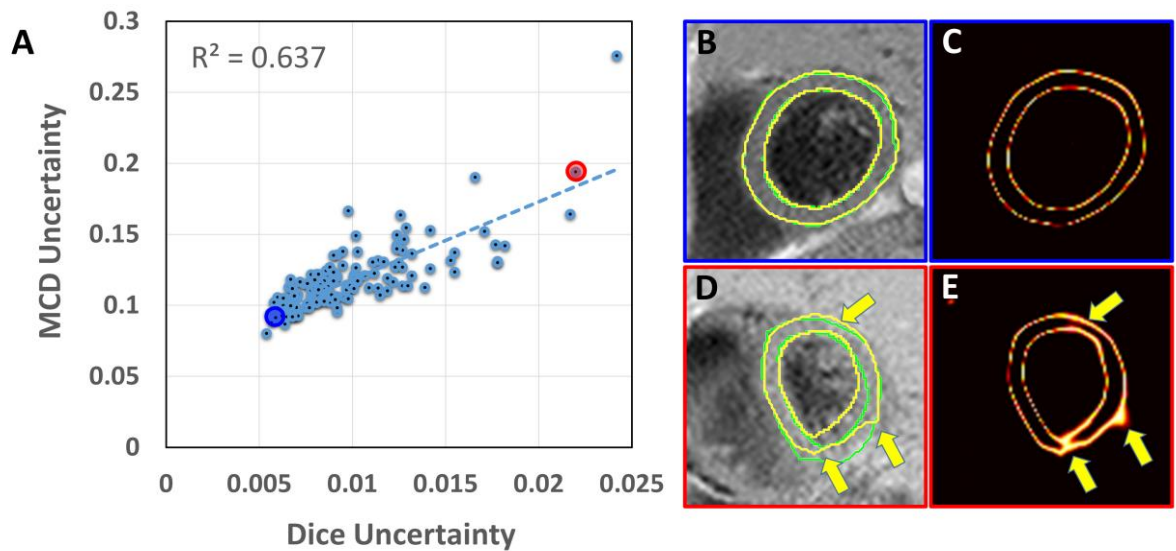

**Figure 5.** Uncertainty metrics. Dice Uncertainty (x-axis) and MCD Uncertainty (y-axis) are correlated ($R^2$ = 0.637) as shown in panel A. Examples of predicted segmentations and uncertainty maps of test cases with low (B and C; blue circle in A) and high (D and E; red circle in A) uncertainty are shown. Manual segmentations are in green and automatic segmentations are in yellow. In the high uncertainty case, the uncertainty map (E) provides the specific spatial locations where the model is most uncertain, which aligns with segmentation errors (D), identified by yellow arrows.

Dice Uncertainty and MCD Uncertainty are in good agreement ($R^2$ = 0.64) as seen in **Figure 5A**. Dice Uncertainty is defined as standard deviation of Dice coefficients calculated from N stochastic predicted segmentations given the ground truth reference segmentation. Higher Dice Uncertainty means the model is less certain of its predictions and experiences higher variability. Calculation of Dice Uncertainty requires the use of the ground truth segmentation, which is often not available at production. Therefore, we proposed MCD Uncertainty as an alternative that is calculated without the use of the ground truth segmentation. The goal of **Figure 5A** is to demonstrate that MCD Uncertainty also represents model uncertainty.

**Figure 5B** and **5C** are the predicted segmentation and uncertainty map of a test case with low global uncertainty score (blue circle in **Figure 5A**). In the test case with low uncertainty, the automatic segmentation (yellow lines) is in good agreement with the manual segmentation (green lines). **Figure 5D** and **5E** are the predicted segmentation and uncertainty map of a test case with high global uncertainty score (red circle in **Figure 5A**). In this case, discrepancies (yellow arrows) can be seen between the automatic segmentation and the manual segmentation. Additionally, the model uncertainty map provides spatial information where the model is most uncertain, as seen in the area indicated by the yellow arrows. MCD Uncertainty is weakly correlated ($R^2 = 0.13$) to physiological noise as seen in **Supporting Information Figure S5**.

As batch size increases, inference time is significantly decreased (**Supporting Information Figure S6A**) without altering the mean prediction and uncertainty measure as seen in **Supporting Information Figure S6B**. The time reduction experienced diminishing return around batch size of 256 with the 12-Gb memory NVIDIA K80 GPU used in this study. With more powerful GPU, the inference time is expected to be further decreased by using larger batch size.

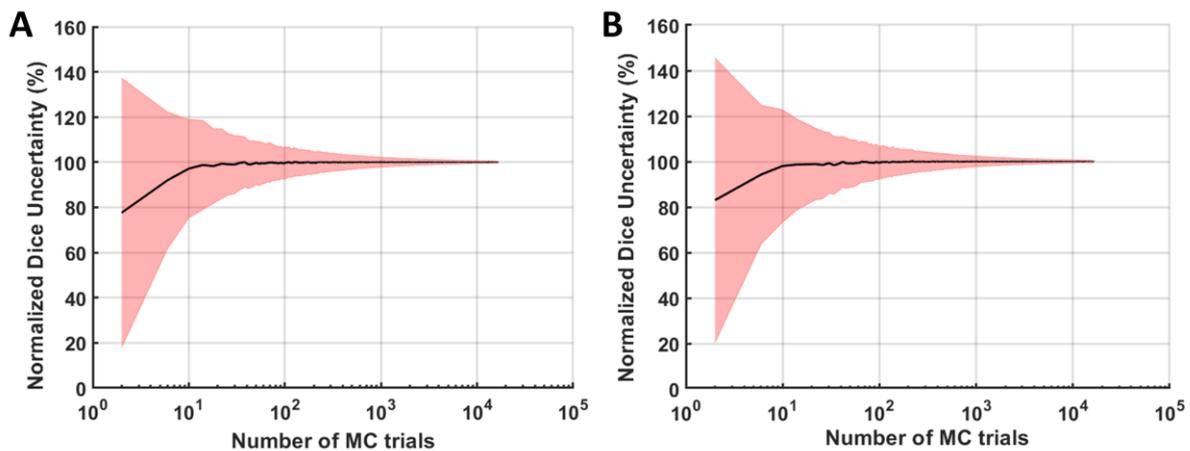

**Figure 6.** Impact of the number of Monte Carlo trials (x-axis) on normalized uncertainty measure (y-axis) from a test case with high uncertainty (A) and another case with low uncertainty (B). Mean (black line) ± one standard deviation (shaded pink) was plotted. It is noted from this figure that more than 100 Monte Carlo trials are needed to achieve accurate uncertainty measures.

As the number of MC trials decreases, the confidence in uncertainty estimation is decreased as seen in **Figure 6**. In this study, we simply chose N = 1115 MC trials for our uncertainty analysis, however, it is worth noting that There is a trade-off between inference time and confidence interval of uncertainty measure and N larger than one hundred can provide accurate uncertainty measure.

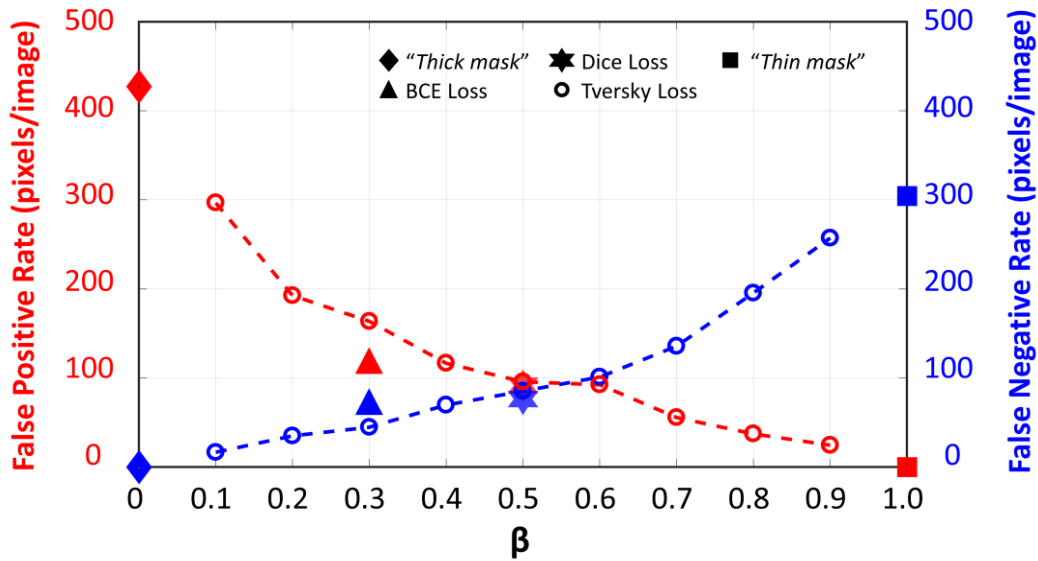

**Figure 7.** Impact of β on the false positive (left y-axis) and false negative (right y-axis) rate. Tversky loss function with different β shows a range of false positive vs. false negative tradeoff. For quantitative cardiovascular magnetic resonance (CMR), β larger than 0.5 may be preferred.

**Adaptability**

**Supporting Information Figure S7** shows an example of a "thin mask" (**A**), a "thick mask" (**C**), and a reference manual mask (**B**). "Thin mask" and "thick mask" data have very similar Dice coefficients, which are 0.80 ± 0.04 and 0.81 ±0.02, respectively. However, the false positive and false negative rates are completely opposite (see **Figure 7**). "Thin mask" had negligible effect to the end-point clinically relevant quantitative MBF while "thick mask" introduced a significant overestimation as seen in **Supporting Information Figure S7D** and **S7E**, respectively. The overestimation is a consequence of partial volume effects (i.e. contamination signal from ventricular blood pools and/or epicardial fat).

**Supporting Information Figure S8A** and **S8B** shows number of false positive pixels subtracted by number of false negative pixels and average Dice coefficient for the entire test set as a function of β. **Figure 7** demonstrates the adaptability of the CNN trained with Tversky loss function. As β increases, false positive and false negative rates are monotonically decreasing and increasing, respectively. By varying β during training, specific false positive vs. false negative tradeoffs can be made. For comparison purposes, false positive and false negative rate from the "thin mask", "thick mask", and CNN models trained with binary cross-entropy, and Dice loss were also shown in **Figure 7**.

**Probability U-Net**

The averaged Dice coefficient from the test set was 0.86 ± 0.04, which is lower than that obtained from MCD U-Net (0.91 ± 0.04). Analysis similar to that in the **Figure 5A** was performed for prob U-Net. The resulting linear regression analysis shows a weaker correlation ($R^2$ = 0.18) between Dice Uncertainty and MCD Uncertainty using prob U-Net. Furthermore, there was no correlation between MCD Uncertainty calculated using MCD U-Net and prob U-Net ($R^2$ = 0.0021).

## DISCUSSION

The contributions of this manuscript are three-fold. First, we demonstrated that it is possible to train a single CNN model to segment control and labeled ASL images, which have substantially different SNR, CNR, and contrast, and whose contrast may vary substantially among images due to heart rate variation. This challenges a common view that a CNN model must be tailored to a specific contrast (with the help of transfer learning). Second, we introduce and evaluate two novel approaches to measure model uncertainty. We denote these "Dice Uncertainty" and "MCD Uncertainty" and calculate them with and without the need for reference manual segmentation, respectively. MCD Uncertainty may be valuable for automatic quality control at production, model improvement via active learning, and/or out-of-distribution detection. Third, we introduce the use of Tversky loss function to adapt the CNN to a specific false positive vs. false negative needs of an application. This is useful for quantitative cardiac MRI since left ventricular myocardium is surrounded by blood pool and epicardial fat, which have distinct physical properties and spin history compared to that of myocardium.

Accuracy

The proposed model achieved good Dice accuracy of 0.91 ± 0.04, similar to those reported in the literature (5,47–49). This was in spite of facing additional challenges compared to cardiac cine imaging, namely 1) lower spatial resolution, 2) lower and inconsistent SNR and blood-myocardium CNR, and 3) SNR and CNR differences between the control and labeled series. Our study also found the quantitative MBF measured using automatic segmentation to be highly correlated with MBF measured using manual segmentation ($R^2 = 0.96$).

Uncertainty

Spatial uncertainty maps were calculated as the standard deviation of all stochastic predictions. A global uncertainty score is needed, in order to perform automatic quality assessment without a human observer. In this study, we introduced and evaluated two simple yet intuitive approaches, denoted "Dice Uncertainty" and "MCD Uncertainty". These were calculated with and without the need for manual segmentation, respectively. We demonstrated that Dice Uncertainty and MCD Uncertainty were in good agreement ($R^2 = 0.64$). Since Dice Uncertainty intuitively represents model uncertainty suggesting that MCD Uncertainty also represents model uncertainty on an input image. MCD Uncertainty does not require the use of reference segmentation therefore it could be used for automatic quality control, for automatic triage of images for active learning, and for out-of-distribution detection.

Uncertainty measures fit several clinical scenarios. In the context of image segmentation, many so-called "automatic" segmentation methods are not "fully automatic" because there lacks a quality metric that could inform the user how confident the prediction is. As a consequence, the human operator often has to review every segmentations generated by an algorithm before they are commenced to the next step in the post-processing pipeline i.e. quantification step. This is still an open challenge for the medical imaging community. This work attempted to address the challenge by introducing MCD Uncertainty measure that may enable a "fully automatic" post-processing pipeline since the human operator may only need to double-check cases with high uncertainty. Human intervention to the high uncertain cases helps to prevent propagations of segmentation error/uncertainty to the perfusion quantification step in the first place avoiding the possibilities that uncertainty/error from segmentation step to confound the integrity of the myocardial perfusion quantification.

There is a time penalty to using MC dropout to compute uncertainty. This study demonstrated that increased batch size significantly decreased inference time without altering mean prediction and uncertainty measure as seen in **Supporting Information Figure S6**. Batch size of 64 or 128 yielded the minimum inference time given the GPU used in this study (NVIDIA K80 GPU). With more powerful GPUs, it is likely that a larger batch size will be feasible and provide shorter inference time. A complimentary approach to reduce inference time is to use a smaller number of MC trials. **Figure 6** demonstrated confidence of uncertainty measure vs. number of MC trials (i.e. inference time). One standard deviation of less than 2% from the expected uncertainty measure was observed with number of MC trials larger than a thousand. In this work, MC trials of 1115 was used to calculate uncertainty metrics, however, accurate uncertainty calculation is achieved with MC trials around a hundred.

There are additional measures of model uncertainty that remain to be explored. Teye et al., demonstrated that Bayesian uncertainty could be estimated from batch normalized deep networks (50). Ayhan et al., demonstrated that model uncertainty could be measured via test-time augmentation (51). Like MC dropout, these methods are efficient and practical, requiring no modification to the existing network architecture. MC dropout may be the most efficient among the three since it does not require any post-processing (i.e. data augmentation or mini-batch preparation). Furthermore, a single input can be replicated and bundled in a mini batch, which is very GPU-efficient. Higher computational capacity allows for larger possible batch sizes, leading to significantly reduced MC inference times.

This work considered only two uncertainty metrics, Dice Uncertainty and MCD Uncertainty, which are both calculated using the variance of the outputs. Neither captures pixel correlations or non-Gaussianity of the output distribution. Further research is needed to investigate metrics that capture pixel correlations and/or consider higher order statistics of the output distribution, such as skewness and kurtosis, which would account for asymmetry and tail extremity, respectively.

Adaptability

Quantitative myocardial measurements can be easily biased if there is even a small contamination with signal from the ventricular blood pool and/or epicardial fat. To improve diagnostic efficacy, false positive may be penalized heavier than false negative during training, leading to a model favorably producing output predictions with a lower false positive rate. In this study, we demonstrated that model adaptability could be achieved using Tversky loss function.

False positive and false negative rate produced by a network depends on the false positive and false negative weighting in the loss function. Dice loss function weights false positive and false negative equally, which is why we observed that false positive and false negative rate from a network trained with Dice loss are very similar, as seen in **Figure 7**. On the other hand, false negative is weighted heavier in binary cross-entropy loss function, leading to a false positive rate higher than the false negative rate. Tversky loss function is a generalized form of Dice loss function with a hyper-parameter $\beta$ that controls the false positive vs. false negative tradeoff. As shown in **Figure 7**, the network trained with Tversky loss function was able to produce ranges of false positive vs. false negative tradeoff. For quantitative CMR, including myocardial ASL, $\beta$ larger than 0.5 may be favorable. The false positive vs. false negative tradeoff of a network could be adapted more broadly in two-dimensional space with two positive hyper-parameters, which independently weight false positive and false negative terms in the Tversky loss function.

Probability U-Net

Prob U-Net is an innovative and attractive concept since it can accurately produce many (even infinitely) plausible hypotheses given an input image. That is especially useful for ambiguous images including medical imaging. One open question for prob U-Net is how to evaluate the quality of each and every plausible hypothesis, to identify the best prediction given an input image. For analyzing results obtained from prob U-Net, we simply used the similar methods as in MCD U-Net to calculate mean prediction, uncertainty map, Dice Uncertainty and MCD Uncertainty, and compared with that calculated from MCD U-Net. We hypothesis that the reason for lower Dice accuracy of prob U-Net (compared to MCD U-Net) is that its designed objective is not best prediction (i.e. prediction with least BCE or Dice loss) but rather to generate many plausible predictions.

Limitations

The primary limitation of this study is that it was performed on a relatively small sample size collected from a single MRI vendor, a single institution, and a single graphical prescription protocol (mid short-axis). This is primarily because ASL-based human myocardial perfusion imaging is an emerging and still experimental technique. Although the sample size is small in absolute terms, this among the largest myocardial ASL datasets from human subjects to date. This study also based on the U-Net model, which has been validated in many medical applications. Dice accuracy on the unseen test set was consistent with those reported in the literature (5,47–49). Model retraining is often required when applied to different datasets, tasks, or applications, even with large training data (5). Therefore, we expect the results from this study can be translated to more variable datasets or other CMR applications through retraining.

## CONCLUSIONS

We demonstrate the feasibility of deep CNN fully-automatic segmentation of the left ventricular myocardium in myocardial ASL perfusion imaging, with good accuracy in terms of Dice coefficients and regional MBF quantification. We introduce two simple yet powerful methods for measuring a global uncertainty score both with and without the use of manual segmentation, termed "Dice Uncertainty" and "MCD Uncertainty", respectively. We also demonstrate the ability to adapt the CNN model to a specific false positive vs. false negative tradeoff using Tversky loss function. These findings are directly relevant to automatic segmentation in quantitative cardiac MRI and are broadly applicable to automatic segmentation problems in diagnostic imaging.

**ACKNOWLEDGEMENTS**

This study was supported by NIH Grant R01HL130494-01A1 and the Whittier Foundation # 0003457-00001

**CONFLICT OF INTEREST**

H.P.D is currently an employee of Canon Medical System USA. The views in this publication do not represent the view of Canon Medical System USA.

**REFERENCES**

1. Kober F, Jao T, Troalen T, Nayak KS. Myocardial arterial spin labeling. J. Cardiovasc. Magn. Reson. 2016;18:22. doi: 10.1186/s12968-016-0235-4.

2. Cerqueira MD, Weissman NJ, Dilsizian V, Jacobs AK, Kaul S, Laskey WK, Pennell DJ, Rumberger J a., Ryan TJ, Verani MS. Standardized Myocardial Segmentation and Nomenclature for Tomographic Imaging of the Heart. Circulation 2002;105:539–542. doi: 10.1081/JCMR-120003946.

3. Kamnitsas K, Ledig C, Newcombe VFJJ, Simpson JP, Kane AD, Menon DK, Rueckert D, Glocker B. Efficient multi-scale 3D CNN with fully connected CRF for accurate brain lesion segmentation. Med. Image Anal. 2017;36:61–78. doi: 10.1016/j.media.2016.10.004.

4. Liu F, Zhou Z, Jang H, Samsonov A, Zhao G, Kijowski R. Deep convolutional neural network and 3D deformable approach for tissue segmentation in musculoskeletal magnetic resonance imaging. Magn. Reson. Med. 2018;79:2379–2391. doi: 10.1002/mrm.26841.

5. Bai W, Sinclair M, Tarroni G, et al. Automated cardiovascular magnetic resonance image analysis with fully convolutional networks. J. Cardiovasc. Magn. Reson. 2018;20:65. doi: 10.1186/s12968-018-0471-x.

6. Milletari F, Navab N, Ahmadi S-A. V-Net: Fully Convolutional Neural Networks for Volumetric Medical Image Segmentation. In 2016 Fourth International Conference on 3D Vision (3DV), pp. 565-571. IEEE, 2016.

7. Puntmann VO, Valbuena S, Hinojar R, Petersen SE, Greenwood JP, Kramer CM, Kwong RY, Mccann GP, Berry C, Nagel E. Society for Cardiovascular Magnetic Resonance (SCMR) expert consensus for CMR imaging endpoints in clinical research: part I-analytical validation and clinical qualification. J. Cardiovasc. Magn. Reson. 2018;20:67. doi: 10.1186/s12968-018-0484-5.

8. Salerno M, Kramer CM. Advances in Parametric Mapping With CMR Imaging. JACC: Cardiovascular Imaging, 6(7), pp.806-822. doi: 10.1016/j.jcmg.2013.05.005.

9. Messroghli DR, Moon JC, Ferreira VM, et al. Clinical recommendations for cardiovascular magnetic resonance mapping of T1, T2, T2* and extracellular volume: A consensus statement by the Society for Cardiovascular Magnetic Resonance (SCMR) endorsed by the European Association for Cardiovascular Imagi. J. Cardiovasc. Magn. Reson. 2017;19:75. doi: 10.1186/s12968-017-0389-8.

10. Nguyen C, Fan Z, Sharif B, He Y, Dharmakumar R, Berman DS, Li D. In Vivo Three-Dimensional High Resolution Cardiac Diffusion-Weighted MRI : A Motion Compensated Diffusion-Prepared Balanced Steady-State Free Precession Approach. Magn. Reson. Med. 2014; 72(5):1257–1267. doi: 10.1002/mrm.25038.

11. Atkins MS, Mackiewich BT. Fully Automatic Segmentation of the Brain in MRI. IEEE Trans. Med. Imaging 1998;17:98–107. doi: 10.1109/42.668699.

12. Petitjean C, Dacher J-N, Petitjean C, Dacher J-N. A review of segmentation methods in short axis cardiac MR images. Med. Image Anal. 2011;15:169–184. doi: https://doi.org/10.1016/j.media.2010.12.004.

13. Gal Y, Islam R, Ghahramani Z. Deep Bayesian Active Learning with Image Data. In Proceedings of the 34th International Conference on Machine Learning-Volume 70, pp. 1183-1192. JMLR. org, 2017.

14. Hinton G. Lecture 10.5 – Dropout: An efficient way to combine neural nets. COURSERA Neural Networks Mach. Learn. 2012:34–41.

15. Srivastava N, Hinton G, Krizhevsky A, Sutskever I, Salakhutdinov R. Dropout: A Simple Way to Prevent Neural Networks from Overfitting. The journal of machine learning research. 2014 Jan 1;15(1):1929-58.

16. Gal Y, Ghahramani Z, Uk ZA, Ghahramani Z, Uk ZA. Dropout as a Bayesian Approximation: Representing Model Uncertainty in Deep Learning. In the international conference on machine learning 2016 Jun 11 (pp. 1050-1059)..

17. Kendall A, Badrinarayanan V, Cipolla R. Bayesian SegNet: Model Uncertainty in Deep Convolutional Encoder-Decoder Architectures for Scene Understanding. arXiv preprint arXiv:1511.02680. 2015 Nov 9.

18. Kendall A, Gal Y. What Uncertainties Do We Need in Bayesian Deep Learning for Computer Vision? In Advances in neural information processing systems 2017 (pp. 5574-5584).

19. Kampffmeyer M, Salberg A-B, Jenssen R. Semantic Segmentation of Small Objects and Modeling of Uncertainty in Urban Remote Sensing Images Using Deep Convolutional Neural Networks. In Proceedings of the IEEE conference on computer vision and pattern recognition workshops 2016 (pp. 1-9).

20. Zhao G, Liu F, Oler JA, Meyerand ME, Kalin NH, Birn RM. Bayesian convolutional neural network based MRI brain extraction on nonhuman primates. Neuroimage 2018;175:32–

44. doi: 10.1016/j.neuroimage.2018.03.065.

21. Kellman P, Hansen MS. T1-mapping in the heart: accuracy and precision. J. Cardiovasc. Magn. Reson. 2014;16:2. doi: 10.1186/1532-429X-16-2.

22. Kellman P, Wilson JR, Xue H, Bandettini WP, Shanbhag SM, Druey KM, Ugander M, Arai AE. Extracellular volume fraction mapping in the myocardium, part 2: initial clinical experience. J. Cardiovasc. Magn. Reson. 2012;14. doi: 10.1186/1532-429X-14-64.

23. Kellman P, Wilson JR, Xue H, Ugander M, Arai AE. Extracellular volume fraction mapping in the myocardium, part 1: evaluation of an automated method. Journal of Cardiovascular Magnetic Resonance. 2012 Dec;14(1):63. doi: 10.1186/1532-429X-14-63.

24. Bulluck H, Rosmini S, Abdel-Gadir A, et al. Automated Extracellular Volume Fraction Mapping Provides Insights Into the Pathophysiology of Left Ventricular Remodeling Post-Reperfused ST-Elevation Myocardial Infarction. J. Am. Heart Assoc. 2016;5:1–12. doi: 10.1161/JAHA.116.003555.

25. Ferreira VM, Wijesurendra RS, Liu A, Greiser A, Casadei B, Robson MD, Neubauer S, Piechnik SK. Systolic ShMOLLI myocardial T1-mapping for improved robustness to partial-volume effects and applications in tachyarrhythmias. J. Cardiovasc. Magn. Reson. 2015;17:1–11. doi: 10.1186/s12968-015-0182-5.

26. Kohl SAA, Romera-Paredes B, Meyer C, De Fauw J, Ledsam JR, Maier-Hein KH, Ali Eslami SM, Jimenez Rezende D, Ronneberger O. A Probabilistic U-Net for Segmentation of Ambiguous Images. In Advances in Neural Information Processing Systems 2018 (pp. 6965-6975).

27. Ronneberger O, Fischer P, Brox T. U-Net: Convolutional Networks for Biomedical Image Segmentation. Int. Conf. Med. image Comput. Comput. Interv. 2015:234–241. doi: 10.1007/978-3-319-24574-4_28.

28. Ioffe S, Szegedy C. Batch Normalization: Accelerating Deep Network Training by Reducing Internal Covariate Shift. arXiv preprint arXiv:1502.03167. 2015 Feb 11.

29. Chollet F, others. Keras. 2015.

30. Abadi M, Barham P, Chen J, et al. TensorFlow: A system for large-scale machine learning. In: 12th USENIX Symposium on Operating Systems Design and Implementation (OSDI 16). ; 2016. pp. 265–283.

31. Do HP, Jao TR, Nayak KS. Myocardial arterial spin labeling perfusion imaging with improved sensitivity. J. Cardiovasc. Magn. Reson. 2014;16:15. doi: 10.1186/1532-429X-

16-15.

32. Do HP, Yoon AJ, Fong MW, Saremi F, Barr ML, Nayak KS. Double-gated myocardial ASL perfusion imaging is robust to heart rate variation. Magn. Reson. Med. 2017;77:1975–1980. doi: 10.1002/mrm.26282.

33. Yoon AJ, Do HP, Cen S, Fong MW, Saremi F, Barr ML, Nayak KS. Assessment of segmental myocardial blood flow and myocardial perfusion reserve by adenosine-stress myocardial arterial spin labeling perfusion imaging. J. Magn. Reson. Imaging 2017;46:413–420. doi: 10.1002/jmri.25604.

34. Kim S-G. Quantification of relative cerebral blood flow change by flow-sensitive alternating inversion recovery (FAIR) technique: Application to functional mapping. Magn. Reson. Med. 1995;34:293–301. doi: 10.1002/mrm.1910340303.

35. Kwong KK, Chesler D a, Weisskoff RM, Donahue KM, Davis TL, Ostergaard L, Campbell T a, Rosen BR. MR perfusion studies with T1-weighted echo planar imaging. Magn. Reson. Med. 1995;34:878–887. doi: 10.1002/mrm.1910340613.

36. Pruessmann KP, Weiger M, Scheidegger MB, Boesiger P. SENSE: Sensitivity encoding for fast MRI. Magn. Reson. Med. 1999;42:952–962. doi: 10.1002/(SICI)1522-2594(199911)42:5<952::AID-MRM16>3.0.CO;2-S.

37. Griswold MA, Jakob PM, Heidemann RM, Nittka M, Jellus V, Wang J, Kiefer B, Haase A. Generalized Autocalibrating Partially Parallel Acquisitions (GRAPPA). Magn. Reson. Med. 2002;47:1202–1210. doi: 10.1002/mrm.10171.

38. Buxton RB, Frank LR, Wong EC, Siewert B, Warach S, Edelman RR. A general kinetic model for quantitative perfusion imaging with arterial spin labeling. Magn. Reson. Med. 1998;40:383–396. doi: 10.1002/mrm.1910400308.

39. Zun Z, Wong EC, Nayak KS. Assessment of myocardial blood flow (MBF) in humans using arterial spin labeling (ASL): feasibility and noise analysis. Magn. Reson. Med. 2009;62:975–983. doi: 10.1002/mrm.22088.

40. Wang G, Li W, Aertsen M, Deprest J, Ourselin S, Vercauteren T. Aleatoric uncertainty estimation with test-time augmentation for medical image segmentation with convolutional neural networks. Neurocomputing. 2019 Apr 21;338:34-45.

41. Milletari F, Navab N, Ahmadi S-A. V-Net: Fully Convolutional Neural Networks for Volumetric Medical Image Segmentation. In 2016 Fourth International Conference on 3D Vision (3DV) 2016 Oct 25 (pp. 565-571). IEEE.

42. Badrinarayanan V, Kendall A, Cipolla R. SegNet: A Deep Convolutional Encoder-Decoder Architecture for Image Segmentation. IEEE Trans. Pattern Anal. Mach. Intell. 2017;39:2481–2495. doi: 10.1109/TPAMI.2016.2644615.

43. Long J, Shelhamer E, Darrell T, Long J, Darrell T. Fully Convolutional Networks for Semantic Segmentation. In Proceedings of the IEEE conference on computer vision and pattern recognition 2015 (pp. 3431-3440).

44. Tversky A. Features of Similarity. Psychol. Rev. 1977;84:327–52.

45. Kingma DP, Ba J, Lei Ba J. Adam: A Method for Stochastic Optimization. arXiv preprint arXiv:1412.6980. 2014 Dec 22.

46. Kohl S. Probabilistic U-Net. github.com 2018: https://github.com/SimonKohl/probabilistic_unet.

47. Zotti C, Luo Z, Humbert O, Lalande A, Jodoin P-M. GridNet with automatic shape prior registration for automatic MRI cardiac segmentation. In International Workshop on Statistical Atlases and Computational Models of the Heart 2017 Sep 10 (pp. 73-81). Springer, Cham.

48. Patravali J, Jain S, Chilamkurthy S. 2D-3D Fully Convolutional Neural Networks for Cardiac MR Segmentation. In International Workshop on Statistical Atlases and Computational Models of the Heart 2017 Sep 10 (pp. 130-139). Springer, Cham.

49. Baumgartner CF, Koch LM, Pollefeys M, Konukoglu E, Given NA. An Exploration of 2D and 3D Deep Learning Techniques for Cardiac MR Image Segmentation. InInternational Workshop on Statistical Atlases and Computational Models of the Heart 2017 Sep 10 (pp. 111-119). Springer, Cham.

50. Teye M, Azizpour H, Smith K. Bayesian Uncertainty Estimation for Batch Normalized Deep Networks. arXiv preprint arXiv:1802.06455. 2018 Feb 18.

51. Ayhan MS, Berens P. Test-time data augmentation for estimation of heteroscedastic aleatoric uncertainty in deep neural networks. 1st Conference on Medical Imaging with Deep Learning (MIDL 2018).

**FIGURE LEGENDS**

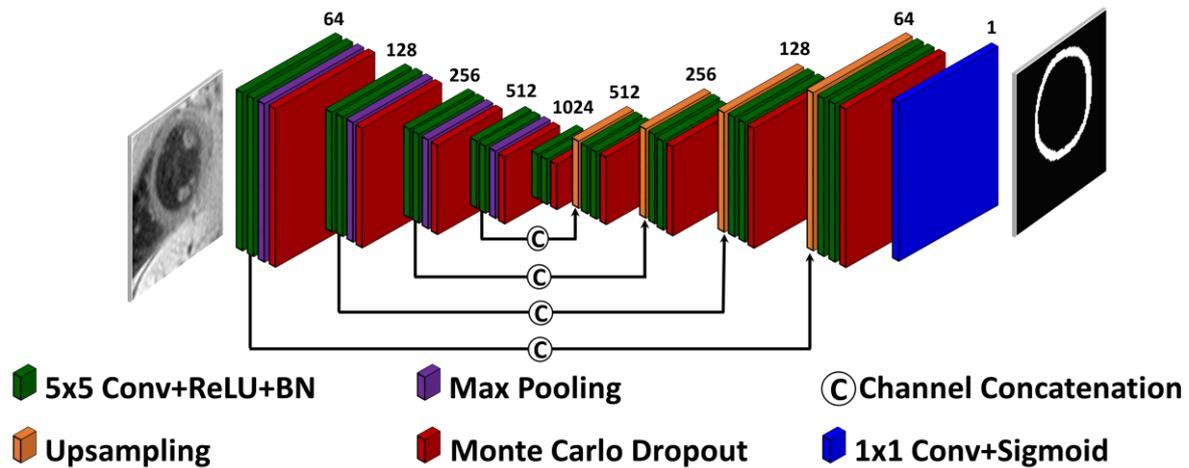

Figure 1. The network architecture of MCD U-Net. The modifications are that (i) 5x5 filter was used for all convolutional layers, (ii) batch normalization was added after each convolutional layer, (iii) dropout rate of 50% was added after each convolutional scale of the down- and up-sampled paths. The number of convolutional kernels is noted on top of the convolutional layers. Conv: convolutional layer; ReLU: rectified linear unit; BN: batch normalization

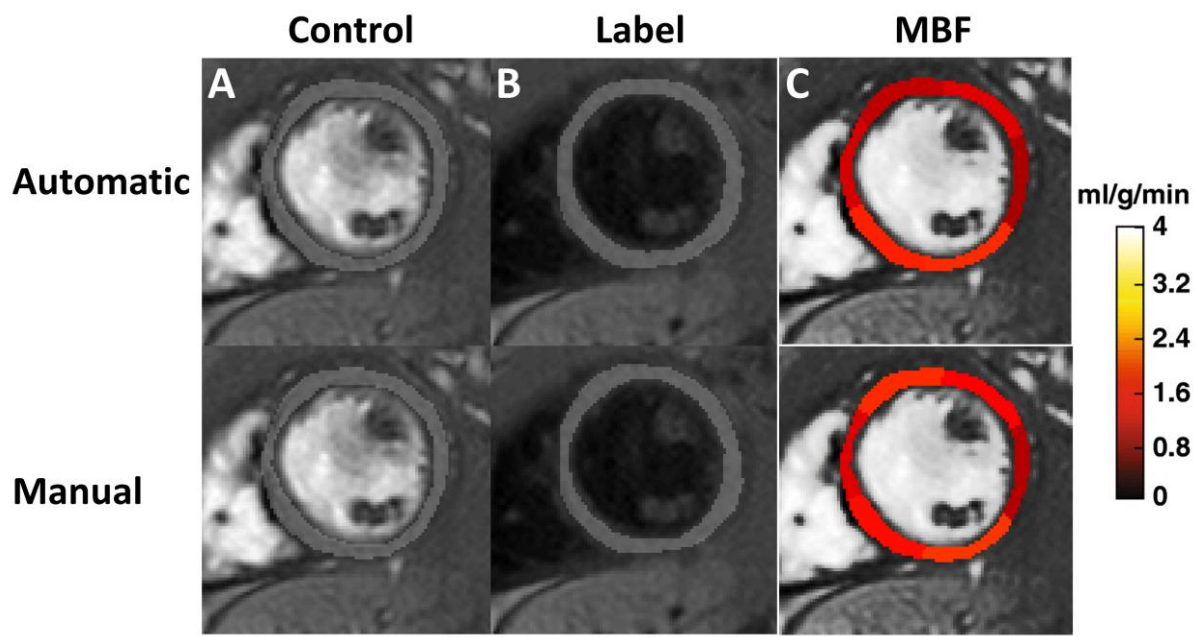

Figure 2. Examples of image quality, segmentation, and myocardial blood flow (MBF) maps. Shown are representative (A) control images, (B) labeled images, and (C) regional MBF maps. Segmentation masks and MBF maps generated by CNN (top row) are comparable to those from manual segmentation (bottom row).

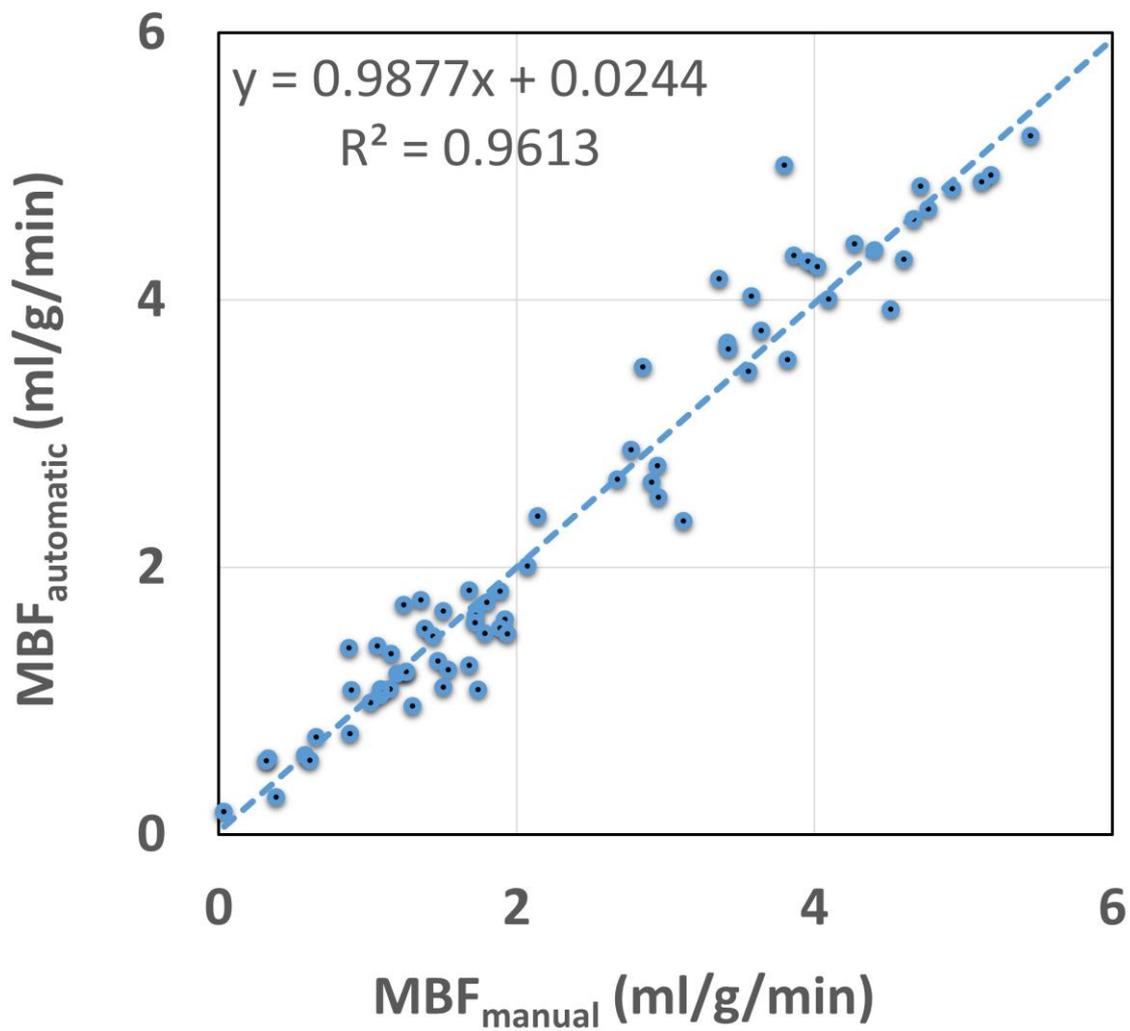

Figure 3. Evaluation of automatic segmentation. Regional myocardial blood flow (MBF) measured using automatic segmentation (y-axis) is highly correlated with that measured using manual segmentation (x-axis) ($R^2$ = 0.96). The concordance correlation coefficient was 0.98.

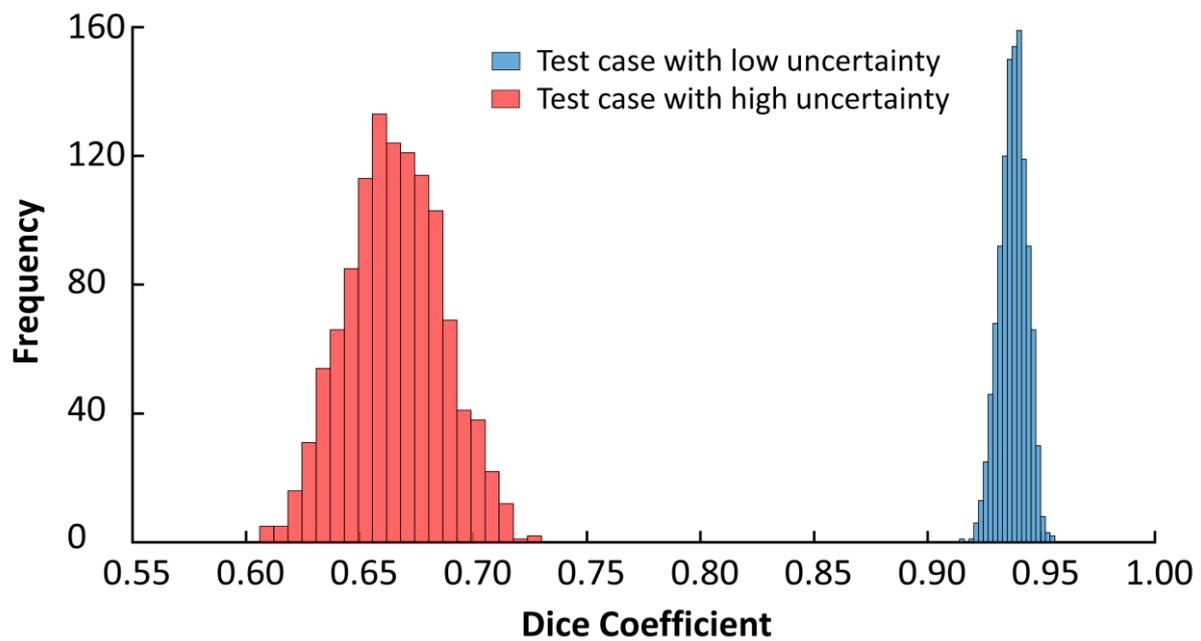

Figure 4. Illustration of the proposed Dice Uncertainty metric. Shown are the Dice coefficient distributions of two test cases after Monte Carlo dropout inference. The distribution is broader in the test case with high uncertainty (red) compared to that with low uncertainty (blue).

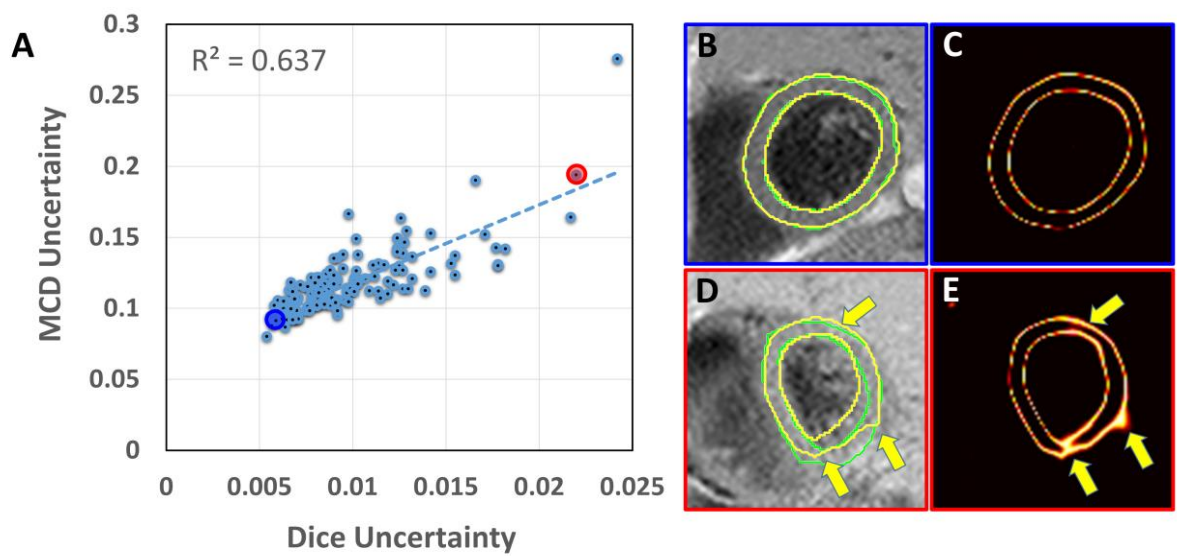

Figure 5.   Uncertainty metrics. Dice Uncertainty (x-axis) and MCD Uncertainty (y-axis) are correlated ($R^2$ = 0.637) as shown in panel A. Examples of predicted segmentations and uncertainty maps of test cases with low (B and C; blue circle in A) and high (D and E; red circle in A) uncertainty are shown. Manual segmentations are in green and automatic segmentations are in yellow. In the high uncertainty case, the uncertainty map (E)

provides the specific spatial locations where the model is most uncertain, which aligns with segmentation errors (D), identified by yellow arrows.

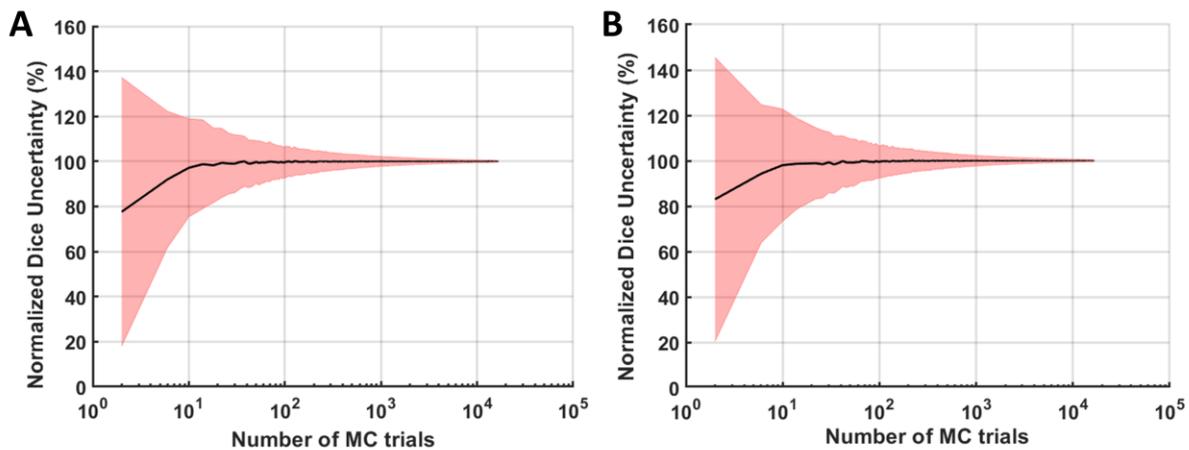

Figure 6. Impact of the number of Monte Carlo trials (x-axis) on normalized uncertainty measure (y-axis) from a test case with high uncertainty (A) and another case with low uncertainty (B). Mean (black line) ± one standard deviation (shaded pink) was plotted. It is noted from this figure that more than 100 Monte Carlo trials are needed to achieve accurate uncertainty measures.

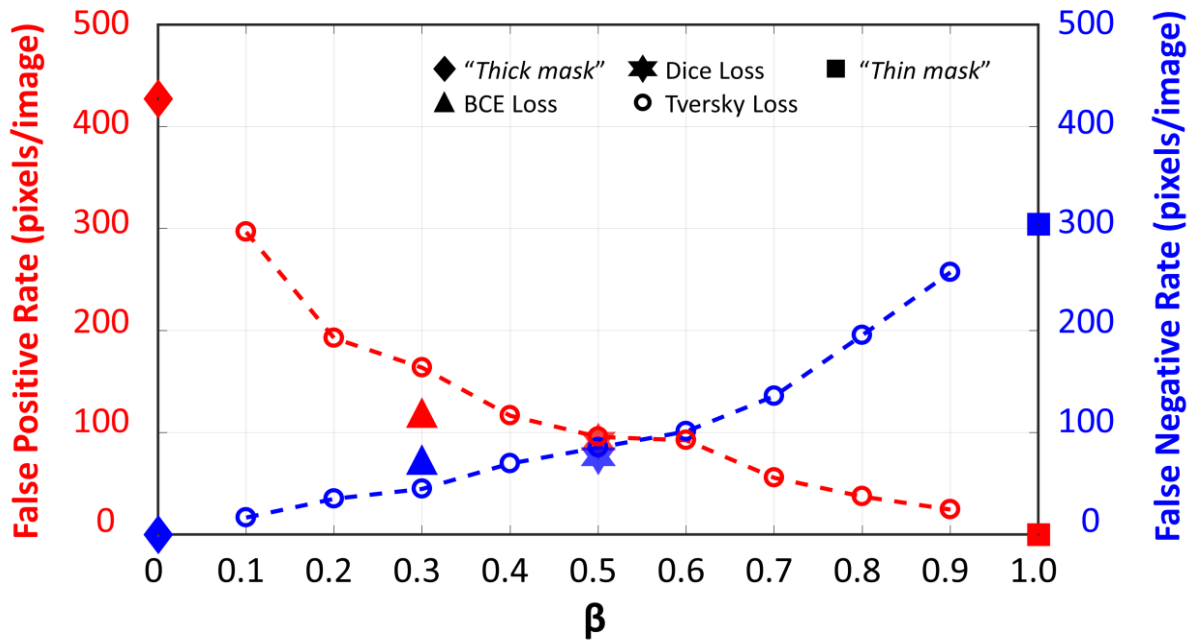

Figure 7. Impact of β on the false positive (left y-axis) and false negative (right y-axis) rate. Tversky loss function with different β shows a range of false positive vs. false negative tradeoff. For quantitative cardiovascular magnetic resonance (CMR), β larger than 0.5 may be preferred.

# LIST OF TABLES

Table 1. Comparison of Dice accuracy from the two training procedures.

|  | Pre-train with Control images and Fine-tune with the Labeled images | | | Train with both Control and Labeled images | |
|---|---|---|---|---|---|
|  | Without Fine-tuning | | Fine-tuning | | |
|  | Control | Labeled | Labeled | Control | Labeled |
| Dice Accuracy | 0.87 ± 0.04 | 0.14 ± 0.08 | 0.86 ± 0.06 | 0.92 ± 0.02 | 0.90 ± 0.04 |

## SUPPORTING INFORMATION

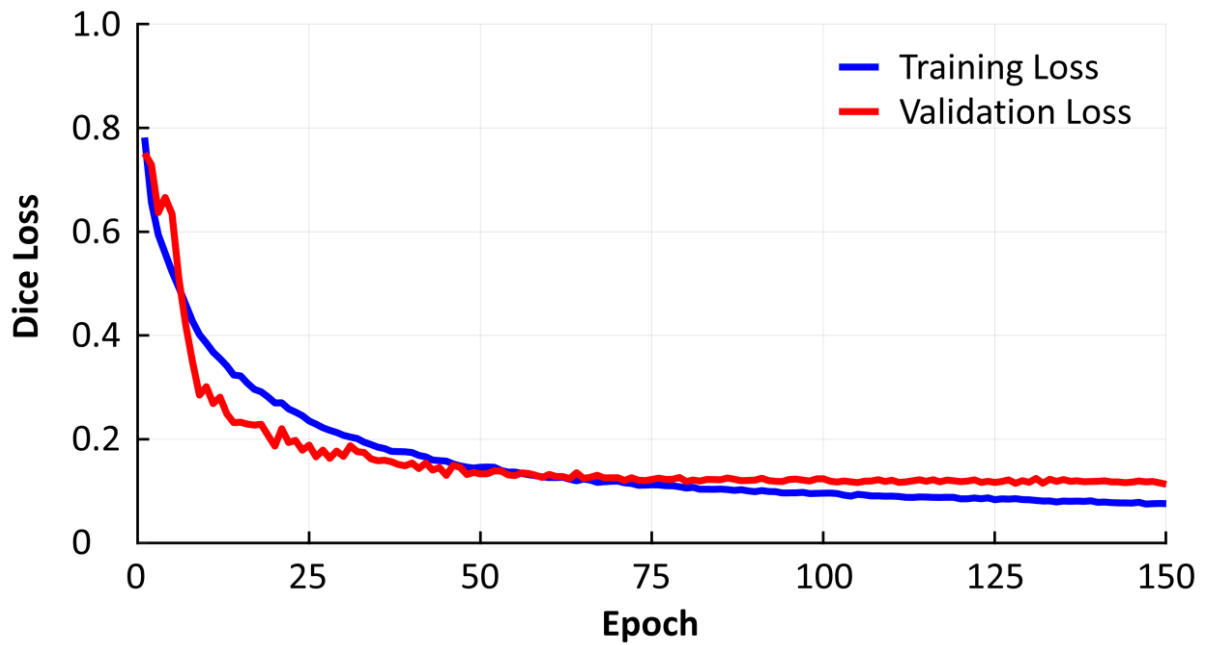

Supporting Information Figure S1: Training and validation loss.

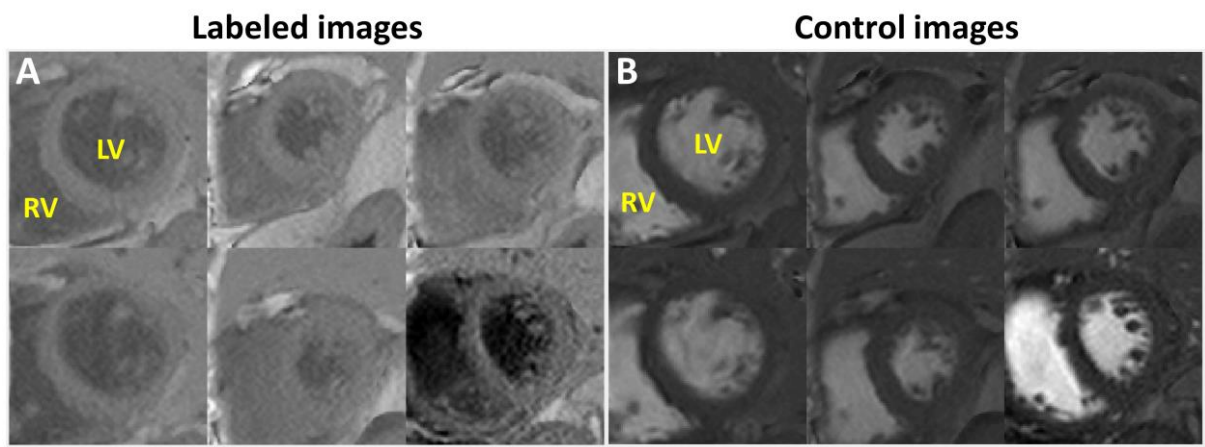

Supporting Information Figure S2: Data characteristics of myocardial arterial spin labeling (ASL). Representative examples of (A) labeled and (B) control images from the test set. Labeled and control images are distinctly different in signal-to-noise ratio (SNR) and contrast-to-noise ratio (CNR). SNR and CNR within each image type are variable due to changes in heart rate during experiments and from patient to patient. RV: right ventricle; LV: left ventricle

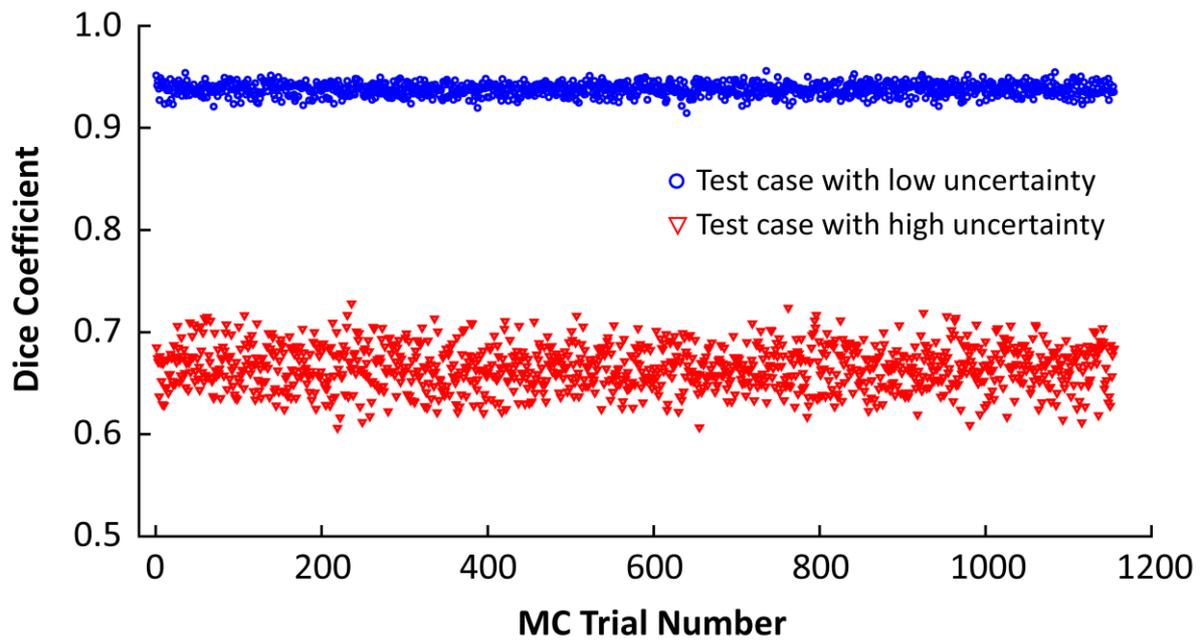

Supporting Information Figure S3: Dice coefficients calculated from N = 1115 stochastic predictions associated with two representative test cases in Figure 4, which have low uncertainty (blue circles) and high uncertainty (red triangles).

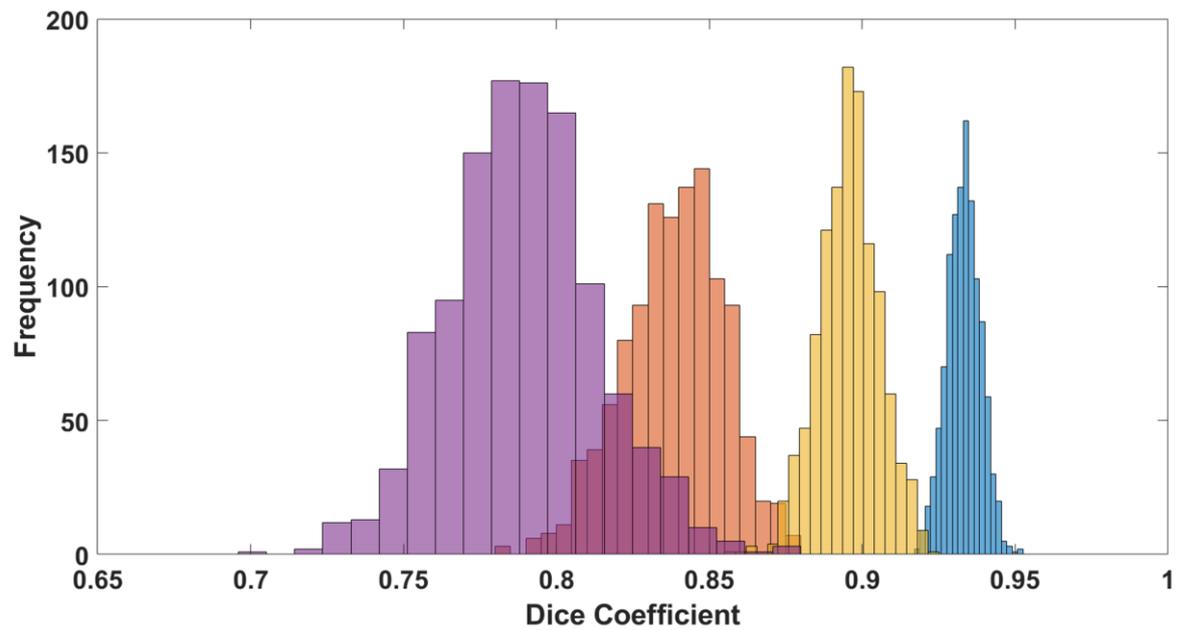

Supporting Information Figure S4: Dice coefficient distributions of four other test cases.

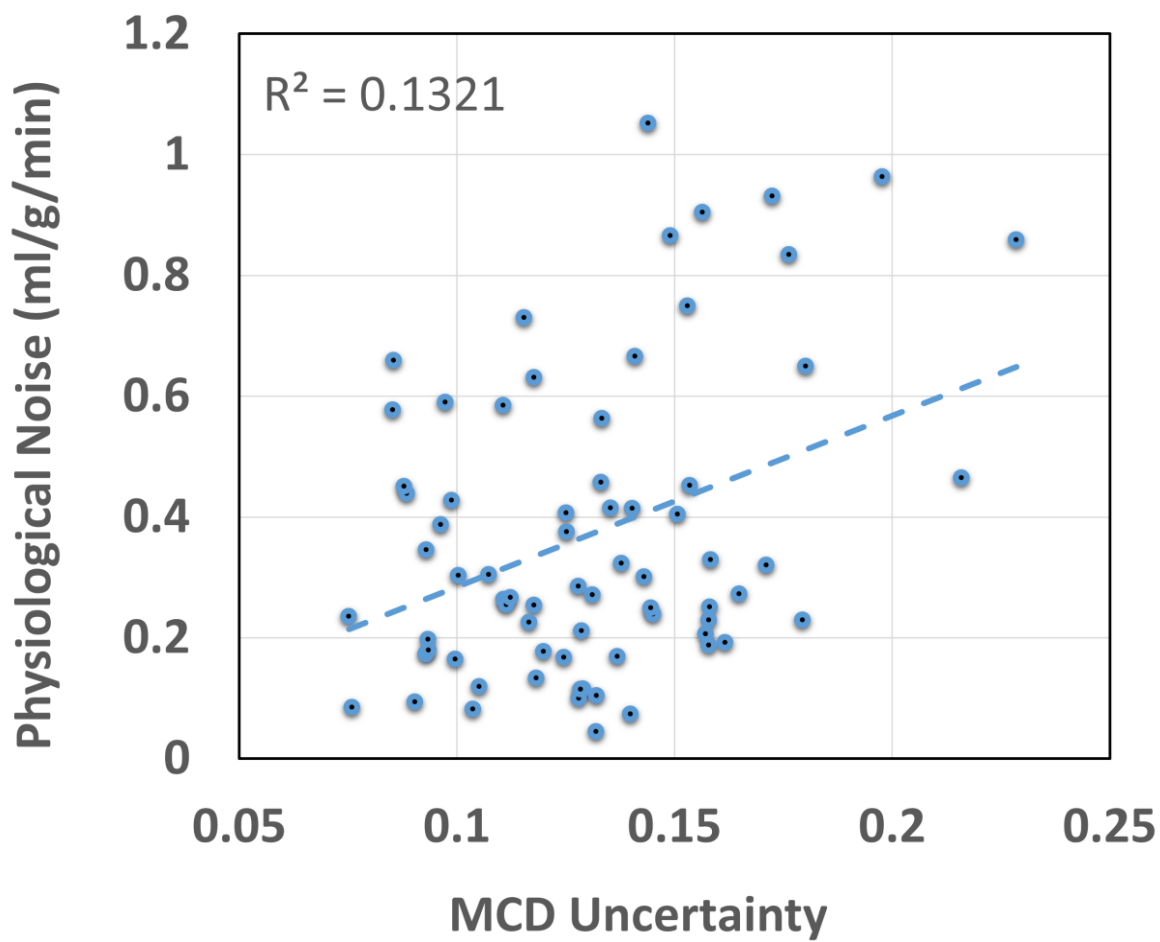

Supporting Information Figure S5: MCD Uncertainty vs. physiological noise (PN). MCD Uncertainty is weakly correlated ($R^2$ = 0.13) to PN.

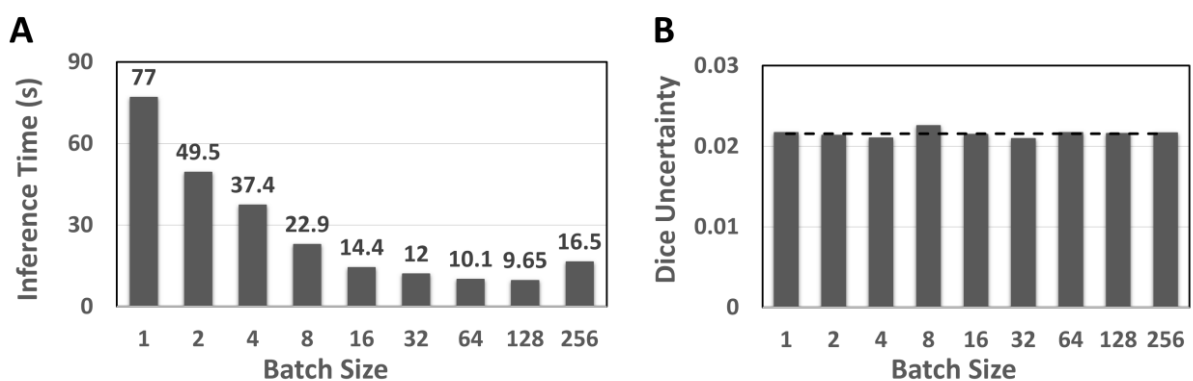

Supporting Information Figure S6: Monte Carlo (MC) dropout inference time with different batch size. Number of MC trials was 1024. Inference time is significantly reduced with increase in batch size (A) without alteration to uncertainty measure (B).

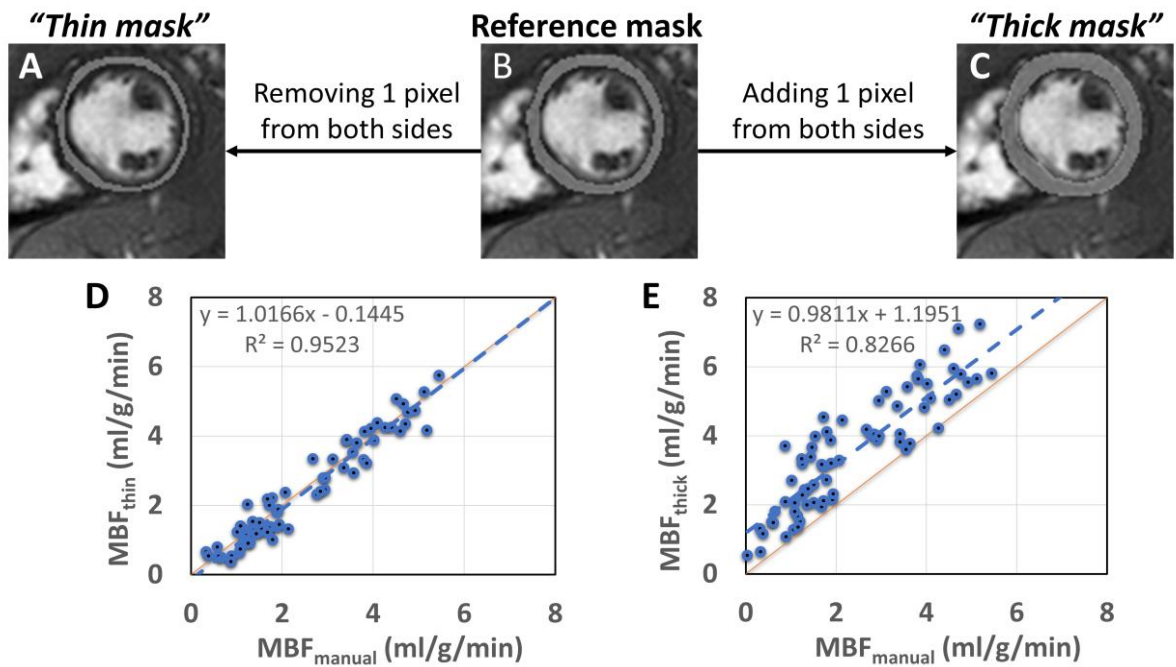

Supporting Information Figure S7: Demonstration of partial volume effects. "Thin mask" (A) and "thick mask" (C) were generated using *bwmorph* Matlab function, which removes and adds one pixel from both sides of the manual mask (B), respectively. Despite distinctly different in the false positive and false negative rate, "thin mask" and "thick mask" data have very similar mean Dice coefficients, which is approximately 0.8. "Thin mask" does not introduce any bias into the end-point quantitative myocardial blood flow (MBF) measure (D). While significant overestimation was observed with "thick mask" (E) due to the partial volume effects.

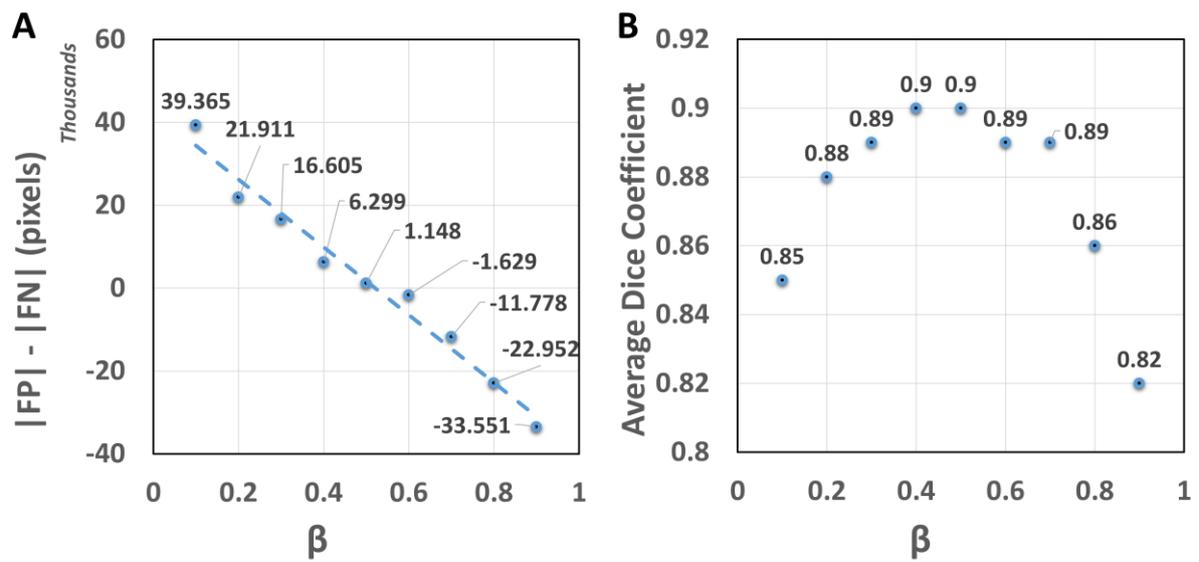

Supporting Information Figure S8: Number of false positive (FP) pixels subtracted by number of false negative (FN) pixels (A) and average Dice coefficient (B) on a test set as a function of β, which is the hyper-parameter in the Tversky loss function.